\documentclass{article}
\usepackage{iclr2017_conference}
\usepackage{times}
\usepackage{latexsym}

\usepackage{latexsym}
\usepackage[utf8]{inputenc} 
\usepackage[T1]{fontenc}   
\usepackage{hyperref}       
\usepackage{url}            
\usepackage{booktabs}       
\usepackage{amsfonts}       
\usepackage{microtype}      
\usepackage[centertags]{amsmath}
\usepackage{amssymb}
\usepackage{graphicx}
\usepackage{tabularx}
\usepackage{multirow}
\usepackage{xspace}

\newcommand{\NN}{\mathcal{N}}
\newcommand{\OO}{\mathcal{O}}

\newcommand{\QQ}{\mathcal{Q}}

\newcommand{\SSS}{\mathcal{S}}

\DeclareMathOperator*{\argmax}{arg\,max}

\newcommand*\equalcontr[1][\value{footnote}]{\footnotemark[#1]}
\def\hevalnewsqaf1{0.694}\xspace
\def\hevalnewsqaem{0.465}\xspace
\def\hevalsquad{0.807}\xspace
\def\hcgap{0.198}\xspace
\def\aibest{0.496}\xspace

\title{NewsQA: A Machine Comprehension Dataset}

\author{Adam Trischler\thanks{These three authors contributed equally.}\qquad\qquad Tong Wang\equalcontr\qquad\qquad Xingdi Yuan\equalcontr\qquad\qquad Justin Harris\\\ \vspace{-2mm}\\{\bf Alessandro Sordoni\qquad\qquad Philip Bachman\qquad\qquad Kaheer Suleman}
\\\ \\ {\tt \{adam.trischler, tong.wang, eric.yuan, justin.harris,}\\ {\tt \ alessandro.sordoni, phil.bachman, k.suleman\}@maluuba.com} \\ Maluuba Research \\ Montr\'{e}al, Qu\'{e}bec, Canada }

\begin{document}

\maketitle

\begin{abstract}
  We present \emph{NewsQA}, a challenging machine comprehension dataset of over 100,000 human-generated question-answer pairs. Crowdworkers supply questions and answers based on a set of over 10,000 news articles from CNN, with answers consisting of spans of text from the corresponding articles. We collect this dataset through a four-stage process designed to solicit exploratory questions that require reasoning. A thorough analysis confirms that \emph{NewsQA} demands abilities beyond simple word matching and recognizing textual entailment. We measure human performance on the dataset and compare it to several strong neural models. The performance gap between humans and machines (\hcgap~in F1) indicates that significant progress can be made on \emph{NewsQA} through future research. The dataset is freely available at \url{https://datasets.maluuba.com/NewsQA}.
\end{abstract}

\section{Introduction}
Almost all human knowledge is recorded in the medium of text. As such, comprehension of written language by machines, at a near-human level, would enable a broad class of artificial intelligence applications. In human students we evaluate reading comprehension by posing questions based on a text passage and then assessing a student's answers. Such comprehension tests are appealing because they are objectively gradable and may measure a range of important abilities, from basic understanding to causal reasoning to inference~\citep{richardson2013}. To teach literacy to machines, the research community has taken a similar approach with machine comprehension (MC).

Recent years have seen the release of a host of MC datasets. Generally, these consist of (document, question, answer) triples to be used in a supervised learning framework. Existing datasets vary in size, difficulty, and collection methodology; however, as pointed out by~\citet{squad}, most suffer from one of two shortcomings: those that are designed explicitly to test comprehension~\citep{richardson2013} are too small for training data-intensive deep learning models, while those that are sufficiently large for deep learning~\citep{hermann2015,hill2015,kadlecdata} are generated synthetically, yielding questions that are not posed in natural language and that may not test comprehension directly~\citep{chenCNN}. More recently, \citet{squad} sought to overcome these deficiencies with their crowdsourced dataset, \emph{SQuAD}.

Here we present a challenging new largescale dataset for machine comprehension: \emph{NewsQA}. \emph{NewsQA} contains 119,633 natural language questions posed by crowdworkers on 12,744 news articles from CNN. Answers to these questions consist of spans of text within the corresponding article highlighted also by crowdworkers. To build \emph{NewsQA} we utilized a four-stage collection process designed to encourage exploratory, curiosity-based questions that reflect human information seeking. CNN articles were chosen as the source material because they have been used in the past~\citep{hermann2015} and, in our view, machine comprehension systems are particularly suited to high-volume, rapidly changing information sources like news.

As~\citet{trischler2016},~\citet{chenCNN}, and others have argued, it is important for datasets to be sufficiently challenging to teach models the abilities we wish them to learn. Thus, in line with~\citet{richardson2013}, our goal with \emph{NewsQA} was to construct a corpus of questions that necessitates reasoning-like behaviors -- for example, synthesis of information across different parts of an article. We designed our collection methodology explicitly to capture such questions.

The challenging characteristics of \emph{NewsQA} that distinguish it from most previous comprehension tasks are as follows:
\begin{enumerate}
\item Answers are spans of arbitrary length within an article, rather than single words or entities.
\item Some questions have no answer in the corresponding article (the \emph{null} span).
\item There are no candidate answers from which to choose.
\item Our collection process encourages lexical and syntactic divergence between questions and answers.
\item A significant proportion of questions requires reasoning beyond simple word- and context-matching (as shown in our analysis).
\end{enumerate}
Some of these characteristics are present also in \emph{SQuAD}, the MC dataset most similar to \emph{NewsQA}. However, we demonstrate through several metrics that \emph{NewsQA} offers a greater challenge to existing models.

In this paper we describe the collection methodology for \emph{NewsQA}, provide a variety of statistics to characterize it and contrast it with previous datasets, and assess its difficulty. In particular, we measure human performance and compare it to that of two strong neural-network baselines. Humans significantly outperform powerful question-answering models. This suggests there is room for improvement through further advances in machine comprehension research.

\section{Related Datasets}
\label{sec:related}
\emph{NewsQA} follows in the tradition of several recent comprehension datasets. These vary in size, difficulty, and collection methodology, and each has its own distinguishing characteristics. We agree with~\citet{kadlecdata} who have said ``models could certainly benefit from as diverse a collection of datasets as possible.'' We discuss this collection below.

\subsection{MCTest}
{\it MCTest}~\citep{richardson2013} is a crowdsourced collection of 660 elementary-level children's stories with associated questions and answers. The stories are fictional, to ensure that the answer must be found in the text itself, and carefully limited to what a young child can understand. Each question comes with a set of 4 candidate answers that range from single words to full explanatory sentences. The questions are designed to require rudimentary reasoning and synthesis of information across sentences, making the dataset quite challenging. This is compounded by the dataset's size, which limits the training of expressive statistical models. Nevertheless, recent comprehension models have performed well on {\it MCTest}~\citep{sachan2015,wangMC}, including a highly structured neural model~\citep{trischler2016}. These models all rely on access to the small set of candidate answers, a crutch that \emph{NewsQA} does not provide.

\subsection{CNN/Daily Mail}
The \emph{CNN/Daily Mail} corpus~\citep{hermann2015} consists of news articles scraped from those outlets with corresponding cloze-style questions. Cloze questions are constructed synthetically by deleting a single entity from abstractive summary points that accompany each article (written presumably by human authors). As such, determining the correct answer relies mostly on recognizing textual entailment between the article and the question. The named entities within an article are identified and anonymized in a preprocessing step and constitute the set of candidate answers; contrast this with \emph{NewsQA} in which answers often include longer phrases and no candidates are given.

Because the cloze process is automatic, it is straightforward to collect a significant amount of data to support deep-learning approaches: \emph{CNN/Daily Mail} contains about 1.4 million question-answer pairs. However,~\citet{chenCNN} demonstrated that the task requires only limited reasoning and, in fact, performance of the strongest models~\citep{kadlec2016,epireader,iaa} nearly matches that of humans.

\subsection{Children's Book Test}
The \emph{Children's Book Test} (\emph{CBT})~\citep{hill2015} was collected using a process similar to that of \emph{CNN/Daily Mail}. Text passages are 20-sentence excerpts from children's books available through Project Gutenberg; questions are generated by deleting a single word in the next ({\it i.e.},~21st) sentence. Consequently, \emph{CBT} evaluates word prediction based on context. It is a comprehension task insofar as comprehension is likely necessary for this prediction, but comprehension may be insufficient and other mechanisms may be more important.

\subsection{BookTest}
\citet{kadlecdata} convincingly argue that, because existing datasets are not large enough, we have yet to reach the full capacity of existing comprehension models. As a remedy they present \emph{BookTest}. This is an extension to the named-entity and common-noun strata of \emph{CBT} that increases their size by over 60 times. \citet{kadlecdata} demonstrate that training on the augmented dataset yields a model~\citep{kadlec2016} that matches human performance on \emph{CBT}. This is impressive and suggests that much is to be gained from more data, but we repeat our concerns about the relevance of story prediction as a comprehension task. We also wish to encourage more efficient learning from less data.

\subsection{SQuAD}
The comprehension dataset most closely related to \emph{NewsQA} is \emph{SQuAD}~\citep{squad}. It consists of natural language questions posed by crowdworkers on paragraphs from high-PageRank Wikipedia articles. As in \emph{NewsQA}, each answer consists of a span of text from the related paragraph and no candidates are provided. Despite the effort of manual labelling, \emph{SQuAD}'s size is significant and amenable to deep learning approaches: 107,785 question-answer pairs based on 536 articles.

Although \emph{SQuAD} is a more realistic and more challenging comprehension task than the other largescale MC datasets, machine performance has rapidly improved towards that of humans in recent months. The \emph{SQuAD} authors measured human accuracy at 0.905 in F1 (we measured human F1 at \hevalsquad~using a different methodology); at the time of writing, the strongest published model to date achieves 0.778 F1~\citep{wang2016multi}. This suggests that new, more difficult alternatives like \emph{NewsQA} could further push the development of more intelligent MC systems.

\section{Collection methodology}
\label{sec:method}
We collected \emph{NewsQA} through a four-stage process:  article curation, question sourcing, answer sourcing, and validation. We also applied a post-processing step with answer agreement consolidation and span merging to enhance the usability of the dataset. These steps are detailed below.

\subsection{Article curation}
We retrieve articles from CNN using the script created by~\citet{hermann2015} for \emph{CNN/Daily Mail}. From the returned set of 90,266 articles, we select 12,744 uniformly at random. These cover a wide range of topics that includes politics, economics, and current events. Articles are partitioned at random into a training set (90\%), a development set (5\%), and a test set (5\%).

\subsection{Question sourcing}
It was important to us to collect challenging questions that could not be answered using straightforward word- or context-matching. Like~\citet{richardson2013} we want to encourage reasoning in comprehension models. We are also interested in questions that, in some sense, model human curiosity and reflect actual human use-cases of information seeking.
Along a similar line, we consider it an important (though as yet overlooked) capacity of a comprehension model to recognize when given information is inadequate, so we are also interested in questions that may not have sufficient evidence in the text.
Our question sourcing stage was designed to solicit questions of this nature, and deliberately separated from the answer sourcing stage for the same reason.

\begin{figure*}
	\centering
	\includegraphics[width=.8\textwidth]{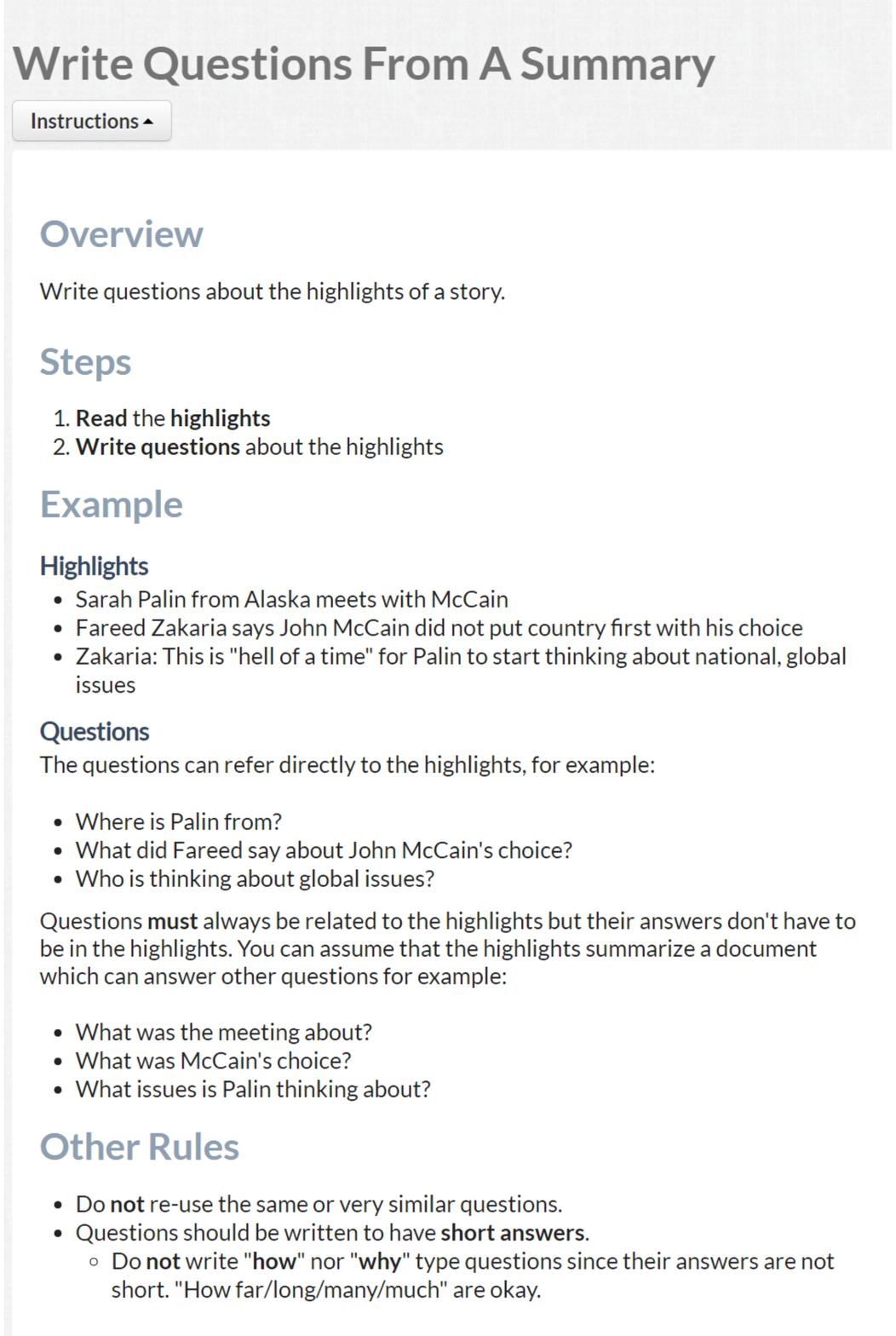}
	\caption{Question sourcing instructions for the crowdworkers.}
	\label{fig:turk-q-source-instructions}
\end{figure*}

{\it Questioners} (a distinct set of crowdworkers) see \emph{only} a news article's headline and its summary points (also available from CNN); they do not see the full article itself. They are asked to formulate a question from this incomplete information. This encourages curiosity about the contents of the full article and prevents questions that are simple reformulations of sentences in the text. It also increases the likelihood of questions whose answers do not exist in the text. We reject questions that have significant word overlap with the summary points to ensure that crowdworkers do not treat the summaries as mini-articles, and further discouraged this in the instructions. During collection each Questioner is solicited for up to three questions about an article. They are provided with positive and negative examples to prompt and guide them (detailed instructions are shown in Figure~\ref{fig:turk-q-source-instructions}).

\subsection{Answer sourcing}
A second set of crowdworkers ({\it Answerers}) provide answers. Although this separation of question and answer increases the overall cognitive load, we hypothesized that unburdening Questioners in this way would encourage more complex questions. Answerers receive a full article along with a crowdsourced question and are tasked with determining the answer. They may also reject the question as nonsensical, or select the {\it null} answer if the article contains insufficient information. Answers are submitted by clicking on and highlighting words in the article, while instructions encourage the set of answer words to consist of a single continuous span (again, we give an example prompt in the Appendix). For each question we solicit answers from multiple crowdworkers (avg. 2.73) with the aim of achieving agreement between at least two Answerers.

\subsection{Validation}
Crowdsourcing is a powerful tool but it is not without peril (collection glitches; uninterested or malicious workers). To obtain a dataset of the highest possible quality we use a validation process that mitigates some of these issues. In validation, a third set of crowdworkers sees the full article, a question, and the set of unique answers to that question. We task these workers with choosing the best answer from the candidate set or rejecting all answers. Each article-question pair is validated by an average of 2.48 crowdworkers. Validation was used on those questions \emph{without} answer-agreement after the previous stage, amounting to 43.2\% of all questions.

\subsection{Answer marking and cleanup}
After validation, 86.0\% of all questions in \emph{NewsQA} have answers agreed upon by at least two separate crowdworkers---either at the initial answer sourcing stage or in the top-answer selection. This improves the dataset's quality.
We choose to include the questions without agreed answers in the corpus also, but they are specially marked. Such questions could be treated as having the \emph{null} answer and used to train models that are aware of poorly posed questions.

As a final cleanup step we combine answer spans that are less than 3 words apart (punctuation is discounted). We find that 5.68\% of answers consist of multiple spans, while 71.3\% of multi-spans are within the 3-word threshold. Looking more closely at the data reveals that the multi-span answers often represent lists. These may present an interesting challenge for comprehension models moving forward.

\section{Data analysis}
\label{sec:anal}
We provide a thorough analysis of \emph{NewsQA} to demonstrate its challenge and its usefulness as a machine comprehension benchmark. The analysis focuses on the types of answers that appear in the dataset and the various forms of reasoning required to solve it.\footnote{Additional statistics are available at \url{https://datasets.maluuba.com/NewsQA/stats}.}

\subsection{Answer types}
\label{sec:answer-types}
Following~\citet{squad}, we categorize answers based on their linguistic type (see Table~\ref{tab:a-type}). This categorization relies on Stanford CoreNLP to generate constituency parses, POS tags, and NER tags for answer spans (see~\citet{squad} for more details). From the table we see that the majority of answers (22.2\%) are common noun phrases. Thereafter, answers are fairly evenly spread among the clause phrase (18.3\%), person (14.8\%), numeric (9.8\%), and other (11.2\%) types. Clearly, answers in \emph{NewsQA} are linguistically diverse.
\begin{table}
	\scriptsize
	\centering
    \caption{The variety of answer types appearing in \emph{NewsQA}, with proportion statistics and examples.}
    \vspace{4pt}
    \begin{tabular}{lcc}
    	\toprule
    	Answer type & Example & Proportion (\%) \\ \midrule
    	Date/Time & March 12, 2008 & 2.9 \\
		  Numeric & 24.3 million & 9.8 \\
		  Person & Ludwig van Beethoven & 14.8 \\
		  Location & Torrance, California & 7.8 \\
		  Other Entity & Pew Hispanic Center & 5.8 \\
		  Common Noun Phr. & federal prosecutors & 22.2 \\
		  Adjective Phr. & 5-hour & 1.9 \\
		  Verb Phr. & suffered minor damage & 1.4 \\
		  Clause Phr. & trampling on human rights & 18.3 \\
		  Prepositional Phr. & in the attack & 3.8 \\
		  Other & nearly half & 11.2 \\
    	\bottomrule
    \end{tabular}
	\label{tab:a-type}
\end{table}

The proportions in Table~\ref{tab:a-type} only account for cases when an answer span exists. The complement of this set comprises questions with an agreed \emph{null} answer (9.5\% of the full corpus) and answers without agreement after validation (4.5\% of the full corpus).

\subsection{Reasoning types}
\label{sec:reasoning-types}
The forms of reasoning required to solve \emph{NewsQA} directly influence the abilities that models will learn from the dataset. We stratified reasoning types using a variation on the taxonomy presented by~\citet{chenCNN} in their analysis of the \emph{CNN/Daily Mail} dataset. Types are as follows, in ascending order of difficulty:
\begin{enumerate}
\item {\bf Word Matching:} Important words in the question exactly match words in the immediate context of an answer span, such that a keyword search algorithm could perform well on this subset.
\item {\bf Paraphrasing:} A single sentence in the article entails or paraphrases the question. Paraphrase recognition may require synonymy and world knowledge.
\item {\bf Inference:} The answer must be inferred from incomplete information in the article or by recognizing conceptual overlap. This typically draws on world knowledge.
\item {\bf Synthesis:} The answer can only be inferred by synthesizing information distributed across multiple sentences.
\item {\bf Ambiguous/Insufficient:} The question has no answer or no unique answer in the article.
\end{enumerate}

For both \emph{NewsQA} and \emph{SQuAD}, we manually labelled 1,000 examples (drawn randomly from the respective development sets) according to these types and compiled the results in Table~\ref{tab:r-type}.
Some examples fall into more than one category, in which case we defaulted to the more challenging type. We can see from the table that word matching, the easiest type, makes up the largest subset in both datasets (32.7\% for \emph{NewsQA} and 39.8\% for \emph{SQuAD}). Paraphrasing constitutes a larger proportion in \emph{SQuAD} than in \emph{NewsQA} (34.3\% vs 27.0\%), possibly a result from the explicit encouragement of lexical variety in \emph{SQuAD} question sourcing. However, \emph{NewsQA} significantly outnumbers \emph{SQuAD} on the distribution of the more difficult forms of reasoning: synthesis and inference make up a combined 33.9\% of the data in contrast to 20.5\% in \emph{SQuAD}.

\begin{table*}[t]
	\scriptsize
	\centering
    \caption{Reasoning mechanisms needed to answer questions. For each we show an example question with the sentence that contains the answer span. Words relevant to the reasoning type are in {\bf bold}. The corresponding proportion in the human-evaluated subset of both \emph{NewsQA} and \emph{SQuAD} (1,000 samples each) is also given.}
    \vspace{4pt}
    \begin{tabularx}{\textwidth}{ l X c c}
    	\toprule
    	\multirow{2}{*}{Reasoning} & \multirow{2}{*}{Example} & \multicolumn{2}{c}{Proportion (\%)} \\
      & & \emph{NewsQA} & \emph{SQuAD} \\
    	\midrule
    	Word Matching &
    	Q: {\bf When were} the {\bf findings} {\bf published}? \newline
    	S: Both sets of research {\bf findings} {\bf were} {\bf published Thursday}... 
    	& 32.7 & 39.8 \\
    	\midrule
		Paraphrasing &
		Q: {\bf Who} is the {\bf struggle between} in Rwanda? \newline
		S: The {\bf struggle} {\bf pits} {\bf ethnic Tutsis}, supported by Rwanda, {\bf against ethnic Hutu}, backed by Congo.
		& 27.0 & 34.3 \\
		\midrule
		Inference &
		Q: {\bf Who} drew {\bf inspiration} from {\bf presidents}? \newline
		S: {\bf Rudy Ruiz} says the lives of US {\bf presidents} can make them {\bf positive role models} for students.
		&  13.2 & 8.6 \\
		\midrule
		Synthesis &
		Q: {\bf Where} is {\bf Brittanee Drexel} from? \newline
		S: The mother of a 17-year-old {\bf Rochester}, {\bf New York} high school student ... says she did not give her daughter permission to go on the trip. {\bf Brittanee} Marie {\bf Drexel}'s mom says...
		& 20.7 & 11.9 \\
		\midrule
		Ambiguous/Insufficient &
		Q: {\bf Whose mother} is {\bf moving} to the White House? \newline
		S: ... {\bf Barack Obama's mother-in-law}, Marian Robinson, will {\bf join} the Obamas at the {\bf family's private quarters} at 1600 Pennsylvania Avenue. [Michelle is never mentioned]
		& 6.4 & 5.4 \\
    	\bottomrule
    \end{tabularx}
	\label{tab:r-type}
\end{table*}

\section{Baseline models}
\label{sec:models}
We test the performance of three comprehension systems on \emph{NewsQA}: human data analysts and two neural models. The first neural model is the match-LSTM (mLSTM) system of~\citet{wangsquad}. The second is a model of our own design that is similar but computationally cheaper. We describe these models below but omit the personal details of our analysts. Implementation details of the models are described in Appendix~\ref{apd:impl-details}.

\subsection{Match-LSTM}
We selected the mLSTM model because it is straightforward to implement and offers strong, though not state-of-the-art, performance on the similar \emph{SQuAD} dataset.
There are three stages involved in the mLSTM. First, LSTM networks encode the document and question (represented by GloVe word embeddings~\citep{pennington2014}) as sequences of hidden states. Second, an mLSTM network~\citep{wang2015snli} compares the document encodings with the question encodings. This network processes the document sequentially and at each token uses an attention mechanism to obtain a weighted vector representation of the question; the weighted combination is concatenated with the encoding of the current token and fed into a standard LSTM. Finally, a Pointer Network uses the hidden states of the mLSTM to select the boundaries of the answer span.
We refer the reader to~\citet{wang2015snli,wangsquad} for full details.

\subsection{The Bilinear Annotation Re-encoding Boundary (BARB) Model}
The match-LSTM is computationally intensive since it computes an attention over the entire question at each document token in the recurrence. To facilitate faster experimentation with \emph{NewsQA} we developed a lighter-weight model (BARB) that achieves similar results on \emph{SQuAD}\footnote{With the configurations for the results reported in Section~\ref{sec:model-perf}, one epoch of training on \emph{NewsQA} takes about 3.9k seconds for \emph{BARB} and 8.1k seconds for \emph{mLSTM}.}. Our model consists of four stages:
\paragraph{Encoding} All words in the document and question are mapped to real-valued vectors using the GloVe embeddings ${\bf W} \in \mathbb{R}^{|V| \times d}$. This yields ${\bf d}_1, \ldots, {\bf d}_n \in \mathbb{R}^d$ and ${\bf q}_1, \ldots, {\bf q}_m \in \mathbb{R}^d$. A bidirectional GRU network~\citep{bahdanau2014} encodes ${\bf d}_i$ into contextual states ${\bf h}_i \in \mathbb{R}^{D_1}$ for the document. The same encoder is applied to ${\bf q}_j$ to derive contextual states ${\bf k}_j \in \mathbb{R}^{D_1}$ for the question.\footnote{A bidirectional GRU concatenates the hidden states of two GRU networks running in opposite directions. Each of these has hidden size $\frac{1}{2}D_1$.}
\paragraph{Bilinear Annotation} Next we compare the document and question encodings using a set of $C$ bilinear transformations,
\begin{equation*}
 {\bf g}_{ij} = {\bf h}_i^T {\bf T}^{[1:C]} {\bf k}_j, \quad {\bf T}^c \in \mathbb{R}^{D_1 \times D_1},~{\bf g}_{ij} \in \mathbb{R}^C,
\end{equation*}
which we use to produce an $(n \times m \times C)$-dimensional tensor of annotation scores, ${\bf G} = [{\bf g}_{ij}]$. We take the maximum over the question-token (second) dimension and call the columns of the resulting matrix ${\bf g}_i \in \mathbb{R}^C$. We use this matrix as an annotation over the document word dimension. In contrast with the more typical multiplicative application of attention vectors, this annotation matrix is concatenated to the encoder RNN input in the re-encoding stage.

\paragraph{Re-encoding} For each document word, the input of the re-encoding RNN (another biGRU) consists of three components: the document encodings $\bf h_i$, the annotation vectors $\bf g_i$, and a binary feature $q_i$ indicating whether the document word appears in the question. The resulting vectors ${\bf f}_i = [{\bf h}_i; {\bf g}_i; q_i]$ are fed into the re-encoding RNN to produce $D_2$-dimensional encodings ${\bf e}_i$ for the boundary-pointing stage.
\paragraph{Boundary pointing} Finally, we search for the boundaries of the answer span using a convolutional network (in a process similar to edge detection). Encodings ${\bf e}_i$ are arranged in matrix ${\bf E} \in \mathbb{R}^{D_2 \times n}$. ${\bf E}$ is convolved with a bank of $n_f$ filters, $\mathbf{F}_k^\ell \in \mathbb{R}^{D_2 \times w}$, where $w$ is the filter width, $k$ indexes the different filters, and $\ell$ indexes the layer of the convolutional network. Each layer has the same number of filters of the same dimensions. We add a bias term and apply a nonlinearity (ReLU) following each convolution, with the result an $(n_f \times n)$-dimensional matrix ${\bf B}_\ell$.

We use two convolutional layers in the boundary-pointing stage. Given ${\bf B}_1$ and ${\bf B}_2$, the answer span's start- and end-location probabilities are computed using $p(s) \propto \exp \left( {\bf  v}_s^T {\bf B}_1 + b_s\right) $ and $p(e) \propto \exp \left( {\bf  v}_e^T {\bf B}_2 + b_e \right)$, respectively. We also concatenate $p(s)$ to the input of the second convolutional layer (along the $n_f$-dimension) so as to condition the end-boundary pointing on the start-boundary. Vectors ${\bf  v}_s$, ${\bf  v}_e \in \mathbb{R}^{n_f}$ and scalars $b_s$, $b_e \in \mathbb{R}$ are trainable parameters.

We also provide an intermediate level of ``guidance'' to the annotation mechanism by first reducing the feature dimension $C$ in $\bf G$ with mean-pooling, then maximizing the softmax probabilities in the resulting ($n$-dimensional) vector corresponding to the answer word positions in each document. This auxiliary task is observed empirically to improve performance.

\section{Experiments\protect\footnote{All experiments in this section use the subset of \emph{NewsQA} dataset with answer agreements (92,549 samples for training, 5,166 for validation, and 5,126 for testing). We leave the challenge of identifying the unanswerable questions for future work.}}
\label{sec:exp}

\subsection{Human evaluation}
We tested four English speakers on a total of 1,000 questions from the \emph{NewsQA} development set. We used four performance measures: F1 and exact match (EM) scores (the same measures used by \emph{SQuAD}), as well as BLEU and CIDEr\footnote{We use \url{https://github.com/tylin/coco-caption} to calculate these two scores.}. BLEU is a precision-based metric popular in machine translation that uses a weighted average of variable length phrase matches ($n$-grams) against the reference sentence~\citep{papineni2002bleu}. CIDEr was designed to correlate better with human judgements of sentence similarity, and uses \emph{tf-idf} scores over $n$-grams~\citep{vedantam2015cider}.

As given in Table~\ref{tab:datasetresults}, humans averaged \hevalnewsqaf1~ F1 on \emph{NewsQA}. The human EM scores are relatively low at \hevalnewsqaem. These lower scores are a reflection of the fact that, particularly in a dataset as complex as \emph{NewsQA}, there are multiple ways to select semantically equivalent answers, {\it e.g.}, ``1996'' versus ``in 1996''. Although these answers are equally correct they would be measured at 0.5 F1 and 0.0 EM. This suggests that simpler automatic metrics are not equal to the task of complex MC evaluation, a problem that has been noted in other domains~\citep{liu2016}. Therefore we also measure according to BLEU and CIDEr: humans score 0.560 and 3.596 on these metrics, respectively.

The original \emph{SQuAD} evaluation of human performance compares distinct answers given by crowdworkers according to EM and F1; for a closer comparison with \emph{NewsQA}, we replicated our human test on the same number of validation data (1,000) with the same humans. We measured human answers against the second group of crowdsourced responses in \emph{SQuAD}'s development set, yielding \hevalsquad~ F1, 0.625 BLEU, and 3.998 CIDEr. Note that the F1 score is close to the top single-model performance of 0.778 achieved in~\cite{wang2016multi}.

We finally compared human performance on the answers that had crowdworker agreement with and without validation, finding a difference of only 1.4 percentage points F1. This suggests our validation stage yields good-quality answers.

\subsection{Model performance}
\label{sec:model-perf}
\begin{table*}[t!]
  \small
  \centering
  \caption{Model performance on \emph{SQuAD} and \emph{NewsQA} datasets. \emph{Random} are taken from~\citet{squad}, and \emph{mLSTM} from~\citet{wangsquad}.}
  \vspace{4pt}
  \begin{tabular}{lcccc}
    \toprule
    \emph{SQuAD} & \multicolumn{2}{c}{Exact Match}   &    \multicolumn{2}{c}{F1}          \\
 	\cmidrule{2-3} \cmidrule{4-5}
    Model & Dev & Test & Dev & Test             \\
    \midrule
    Random & 0.11 & 0.13 & 0.41 & 0.43 \\
    mLSTM & 0.591 & 0.595 & 0.700 & 0.703 \\
    BARB & 0.591 & - & 0.709 & - \\
    \bottomrule
  \end{tabular}
  \quad
  \begin{tabular}{lcccc}
    \toprule
    \emph{NewsQA} & \multicolumn{2}{c}{Exact Match}   &    \multicolumn{2}{c}{F1}          \\
 	\cmidrule{2-3} \cmidrule{4-5}
    Model & Dev & Test & Dev & Test             \\
    \midrule
    Random & 0.00 & 0.00 & 0.30 & 0.30 \\
    mLSTM & 0.344 & 0.349 & 0.496 & 0.500 \\
    BARB & 0.361 & 0.341 & \aibest & 0.482 \\
    \bottomrule
  \end{tabular}
  \label{tab:datasetresults}
\end{table*}

\begin{table}[t!]
  \small
  \centering
  \caption{Human performance on \emph{SQuAD} and \emph{NewsQA} datasets. The first row is taken from~\citet{squad}, and the last two rows correspond to machine performance (BARB) on the human-evaluated subsets.}
  \vspace{4pt}
  \begin{tabular}{lcccc}
    \toprule
    Dataset & Exact Match & F1 & BLEU & CIDEr\\
    \midrule
    \emph{SQuAD} & 0.803 & 0.905 & - & -\\
    \emph{SQuAD} (ours) & 0.650 & \hevalsquad & 0.625 & 3.998\\
    \emph{NewsQA} & \hevalnewsqaem & \hevalnewsqaf1 & 0.560 & 3.596\\
    \midrule
    \emph{SQuAD}$_\mathrm{BARB}$ & 0.553 & 0.685 & 0.366 & 2.845\\
    \emph{NewsQA}$_\mathrm{BARB}$ & 0.340 & 0.501 & 0.081 & 2.431\\
    \bottomrule
  \end{tabular}
  \label{tab:datasetresults}
\end{table}

Performance of the baseline models and humans is measured by EM and F1 with the official evaluation script from \emph{SQuAD} and listed in Table~\ref{tab:datasetresults}. We supplement these with BLEU and CIDEr measures on the 1,000 human-annotated dev questions. Unless otherwise stated, hyperparameters are determined by \texttt{hyperopt} (Appendix~\ref{apd:impl-details}). The gap between human and machine performance on \emph{NewsQA} is a striking \hcgap~points F1 --- much larger than the gap on \emph{SQuAD} (0.098) under the same human evaluation scheme. The gaps suggest a large margin for improvement with machine comprehension methods.

\begin{figure*}
	\centering
	\includegraphics[width=.49\textwidth]{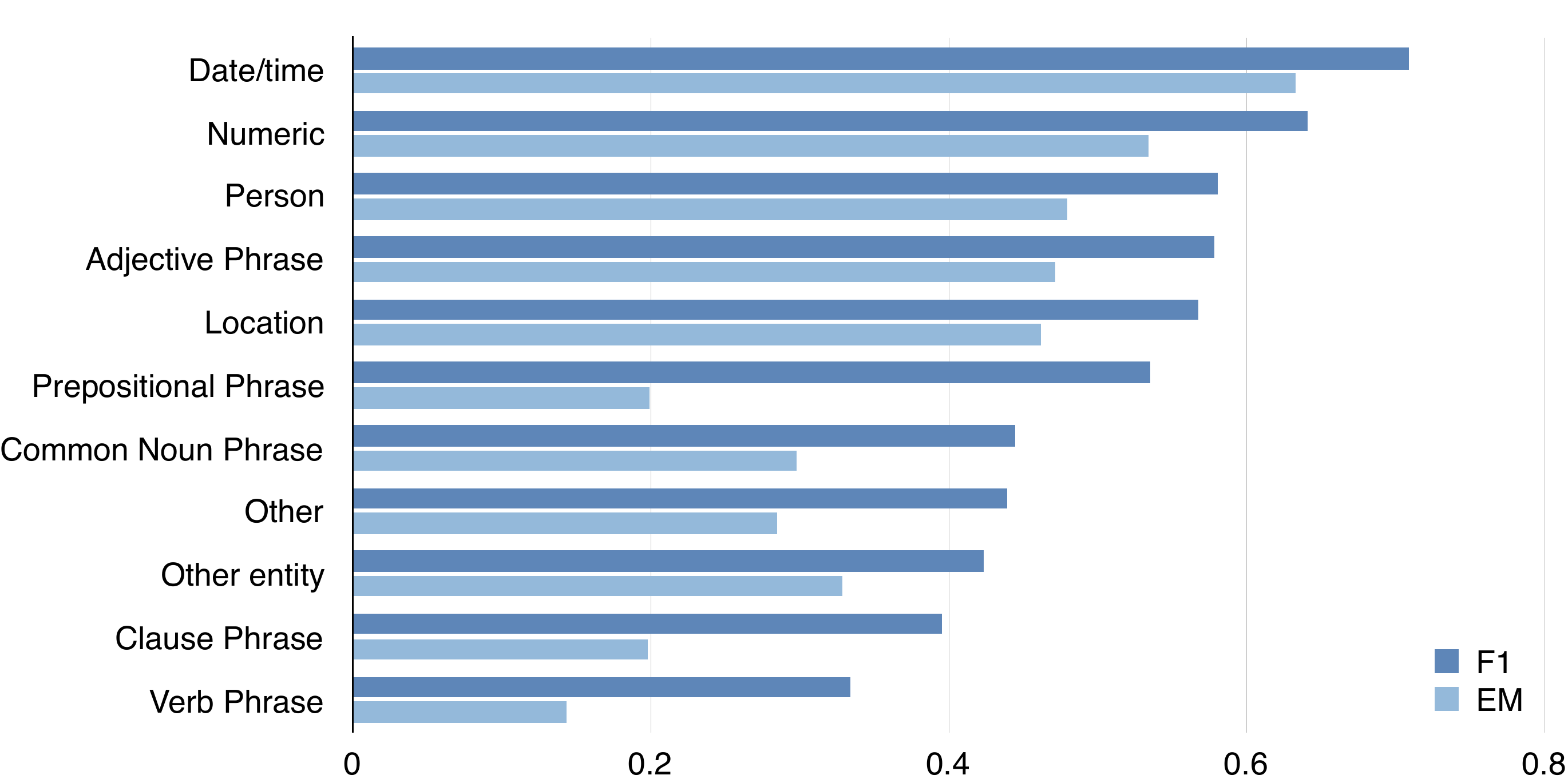}
	\hfill
	\includegraphics[width=.49\textwidth]{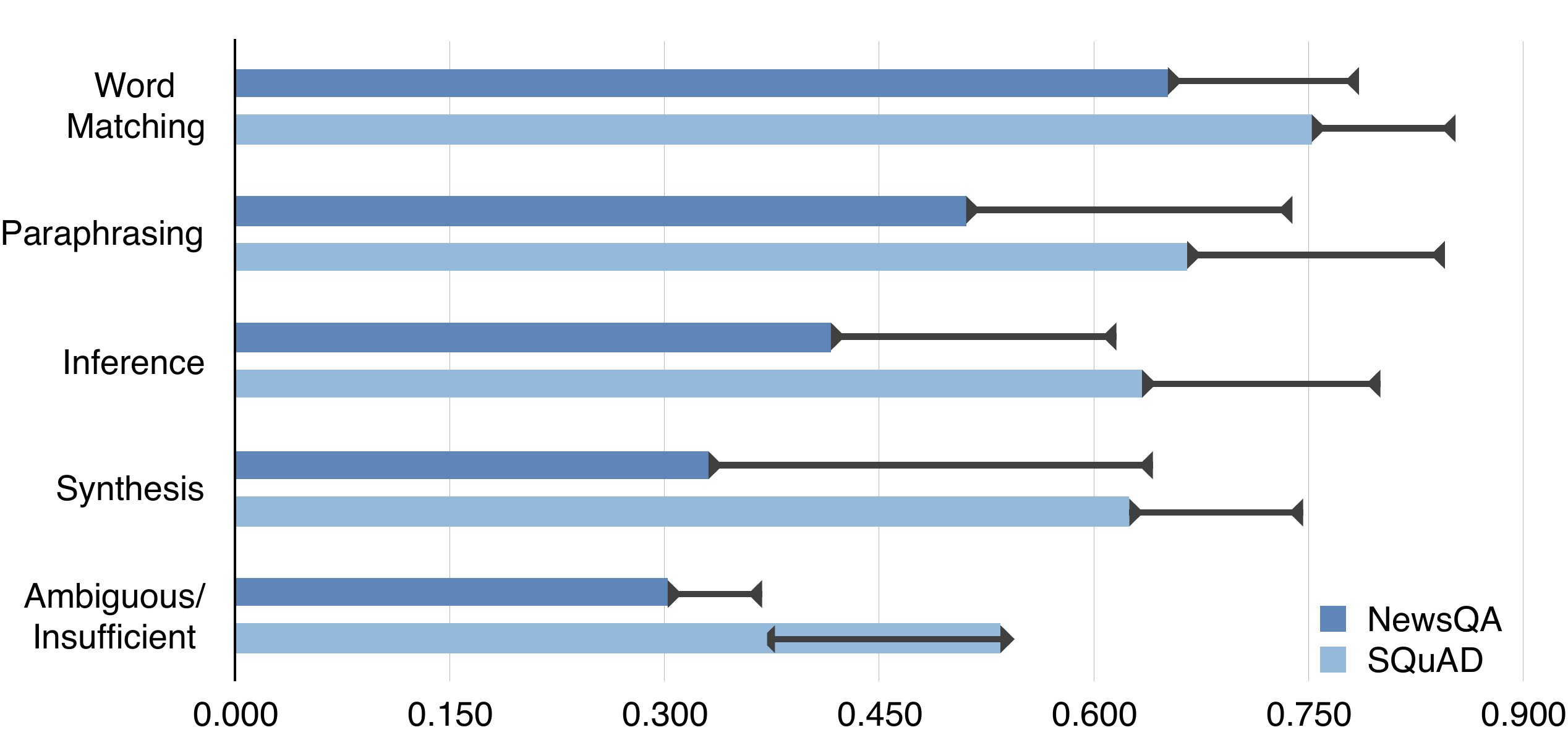}
  \caption{\emph{Left}: BARB performance (F1 and EM) stratified by answer type on the full development set of \emph{NewsQA}.\hspace{4mm}\emph{Right}: BARB performance (F1) stratified by reasoning type on the human-assessed subset on both \emph{NewsQA} and \emph{SQuAD}. Error bars indicate performance differences between BARB and human annotators.}
	\label{fig:stratification-at-rt}
\end{figure*}

Figure~\ref{fig:stratification-at-rt} stratifies model (BARB) performance according to answer type (left) and reasoning type (right) as defined in Sections~\ref{sec:answer-types} and~\ref{sec:reasoning-types}, respectively. The answer-type stratification suggests that the model is better at pointing to named entities compared to other types of answers. The reasoning-type stratification, on the other hand, shows that questions requiring \emph{inference} and \emph{synthesis} are, not surprisingly, more difficult for the model. Consistent with observations in Table~\ref{tab:datasetresults}, stratified performance on \emph{NewsQA} is significantly lower than on \emph{SQuAD}. The difference is smallest on word matching and largest on synthesis. We postulate that the longer stories in \emph{NewsQA} make synthesizing information from separate sentences more difficult, since the relevant sentences may be farther apart. This requires the model to track longer-term dependencies. It is also interesting to observe that on \emph{SQuAD}, BARB outperforms human annotators in answering ambiguous questions or those with incomplete information.

\subsection{Sentence-level scoring}
We propose a simple sentence-level subtask as an additional quantitative demonstration of the relative difficulty of \emph{NewsQA}. Given a document and a question, the goal is to find the sentence containing the answer span. We hypothesize that simple techniques like word-matching are inadequate to this task owing to the more involved reasoning required by \emph{NewsQA}.

We employ a technique that resembles inverse document frequency (\emph{idf}), which we call inverse sentence frequency (\emph{isf}). Given a sentence $\SSS_i$ from an article and its corresponding question $\QQ$, the \emph{isf} score is given by the sum of the \emph{idf} scores of the words common to $\SSS_i$ and $\QQ$ (each sentence is treated as a document for the \emph{idf} computation). The sentence with the highest \emph{isf} is taken as the answer sentence $\SSS_*$, that is,
\[ \SSS_* = \argmax_i \sum_{w \in \SSS_i \cap \QQ} \mathit{isf}(w) .\]

\begin{table}[t!]
  \small
  \centering
  \caption{Sentence-level accuracy on artificially-lengthened \emph{SQuAD} documents.}
  \vspace{4pt}
  \begin{tabularx}{.5\textwidth}{rXXXXXr}
    \toprule
    & \multicolumn{5}{c}{\emph{SQuAD}} & \emph{NewsQA} \\
    \midrule
    \# documents & 1 & 3 & 5 & 7 & 9 & 1 \\
    Avg \# sentences & 4.9 & 14.3 & 23.2 & 31.8 & 40.3 & 30.7\\
    \emph{isf} & 79.6 & 74.9 & 73.0 & 72.3 & 71.0 & 35.4\\
    \bottomrule
  \end{tabularx}
  \label{tab:lengthy-squad}
\end{table}

The \emph{isf} method achieves an impressive 79.4\% sentence-level accuracy on \emph{SQuAD}'s development set but only 35.4\% accuracy on \emph{NewsQA}'s development set, highlighting the comparative difficulty of the latter. To eliminate the difference in article length as a possible cause of the performance gap, we also artificially increased the article lengths in \emph{SQuAD} by concatenating adjacent \emph{SQuAD} articles {\it from the same Wikipedia article}. Accuracy decreases as expected with the increased \emph{SQuAD} article length, yet remains significantly higher than on \emph{NewsQA} with comparable or even greater article length (see Table~\ref{tab:lengthy-squad}).

\section{Conclusion}
\label{sec:conc}
We have introduced a challenging new comprehension dataset: \emph{NewsQA}. We collected the 100,000+ examples of \emph{NewsQA} using teams of crowdworkers, who variously read CNN articles or highlights, posed questions about them, and determined answers. Our methodology yields diverse answer types and a significant proportion of questions that require some reasoning ability to solve. This makes the corpus challenging, as confirmed by the large performance gap between humans and deep neural models (\hcgap~F1, 0.479 BLEU, 1.165 CIDEr).
By its size and complexity, \emph{NewsQA} makes a significant extension to the existing body of comprehension datasets. We hope that our corpus will spur further advances in machine comprehension and guide the development of literate artificial intelligence.

\section*{Acknowledgments}
The authors would like to thank \c{C}a\u{g}lar G\"{u}l\c{c}ehre, Sandeep Subramanian and Saizheng Zhang for helpful discussions.

\bibliography{newsqa}
\bibliographystyle{iclr2017_conference}

\newpage
\section*{Appendices}
\appendix
\section{Implementation details}
\label{apd:impl-details}
Both mLSTM and BARB are implemented with the Keras framework \citep{keras} using the Theano \citep{theano10} backend. Word embeddings are initialized using GloVe vectors \citep{pennington2014} pre-trained on the 840-billion \emph{Common Crawl} corpus. The word embeddings are not updated during training. Embeddings for out-of-vocabulary words are initialized with zero.

For both models, the training objective is to maximize the log likelihood of the boundary pointers. Optimization is performed using stochastic gradient descent (with a batch-size of 32) with the ADAM optimizer \citep{kingma2014}. The initial learning rate is 0.003 for mLSTM and 0.0005 for BARB. The learning rate is decayed by a factor of 0.7 if validation loss does not decrease at the end of each epoch. Gradient clipping \citep{pascanu2013difficulty} is applied with a threshold of 5.

Parameter tuning is performed on both models using \texttt{hyperopt}\footnote{\url{https://github.com/hyperopt/hyperopt}}. For each model, configurations for the best observed performance are as follows:

\textbf{mLSTM}

Both the pre-processing layer and the answer-pointing layer use bi-directional RNN with a hidden size of 192. These settings are consistent with those used by \citet{wangsquad}.

Model parameters are initialized with either the normal distribution ($\NN(0,0.05)$) or the orthogonal initialization ($\OO$, \citealt{saxe2013exact}) in Keras. All weight matrices in the LSTMs are initialized with $\OO$. In the Match-LSTM layer, $W^q$, $W^p$, and $W^r$ are initialized with $\OO$, $b^p$ and $w$ are initialized with $\NN$, and $b$ is initialized as 1.

In the answer-pointing layer, $V$ and $W^a$ are initialized with $\OO$, $b^a$ and $v$ are initialized with $\NN$, and $c$ is initialized as 1.

\textbf{BARB}

For BARB, the following hyperparameters are used on both \emph{SQuAD} and \emph{NewsQA}: $d=300$, $D_1=128$, $C=64$, $D_2=256$, $w=3$, and $n_f=128$. Weight matrices in the GRU, the bilinear models, as well as the boundary decoder (${\bf v_s}$ and ${\bf v_e}$) are initialized with $\OO$. The filter weights in the boundary decoder are initialized with \emph{glorot\_uniform} (\citealt{glorot2010understanding}, default in Keras). The bilinear biases are initialized with $\NN$, and the boundary decoder biases are initialized with 0.

\section{Data collection user interface}
Here we present the user interfaces used in question sourcing, answer sourcing, and question/answer validation.

\begin{figure*}
	\centering
	\includegraphics[width=.8\textwidth]{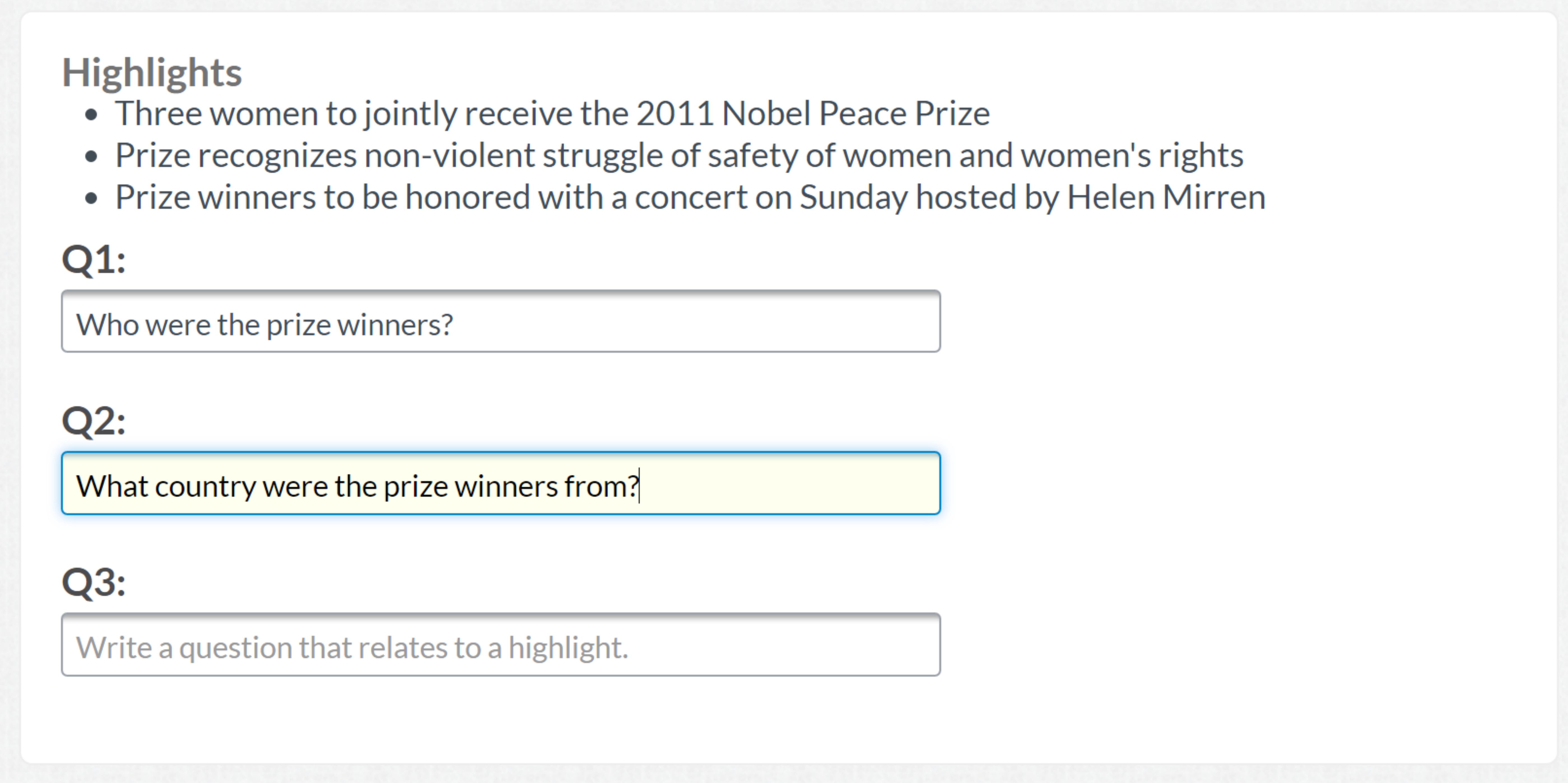}
	\includegraphics[width=.8\textwidth]{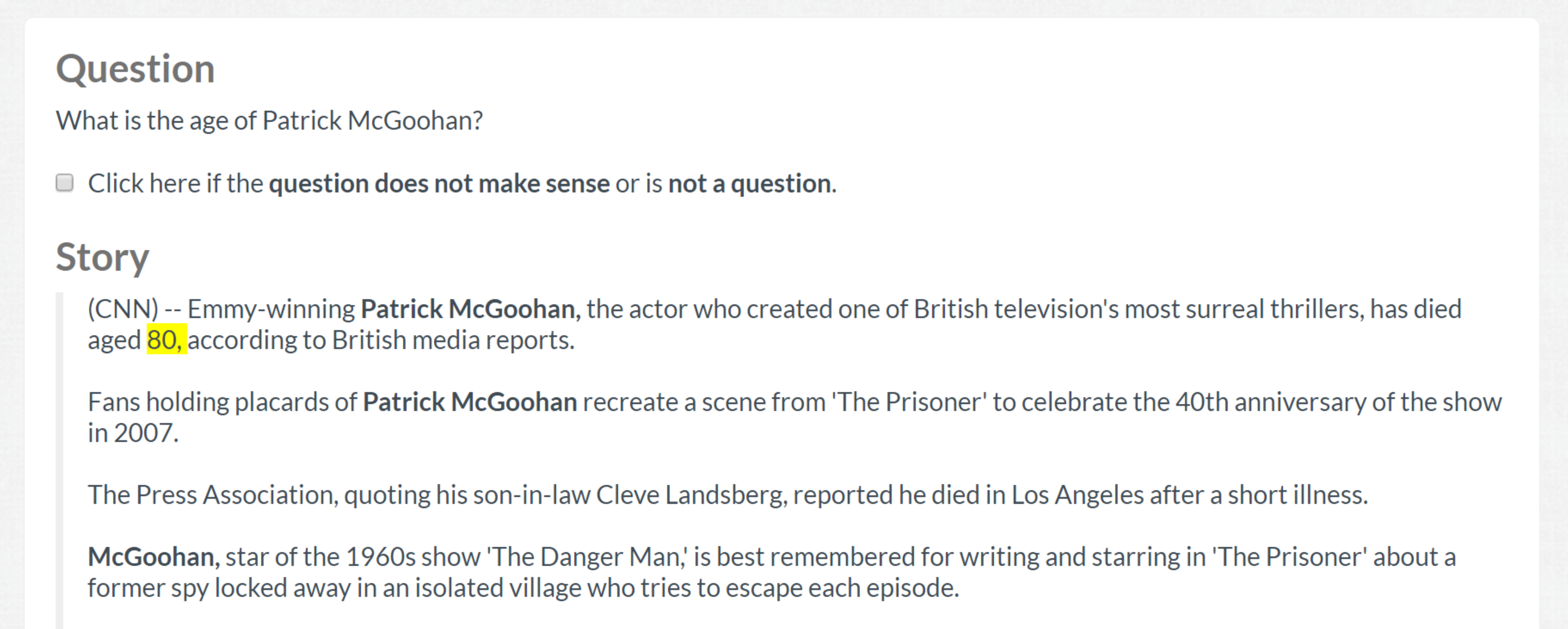}
	\includegraphics[width=.8\textwidth]{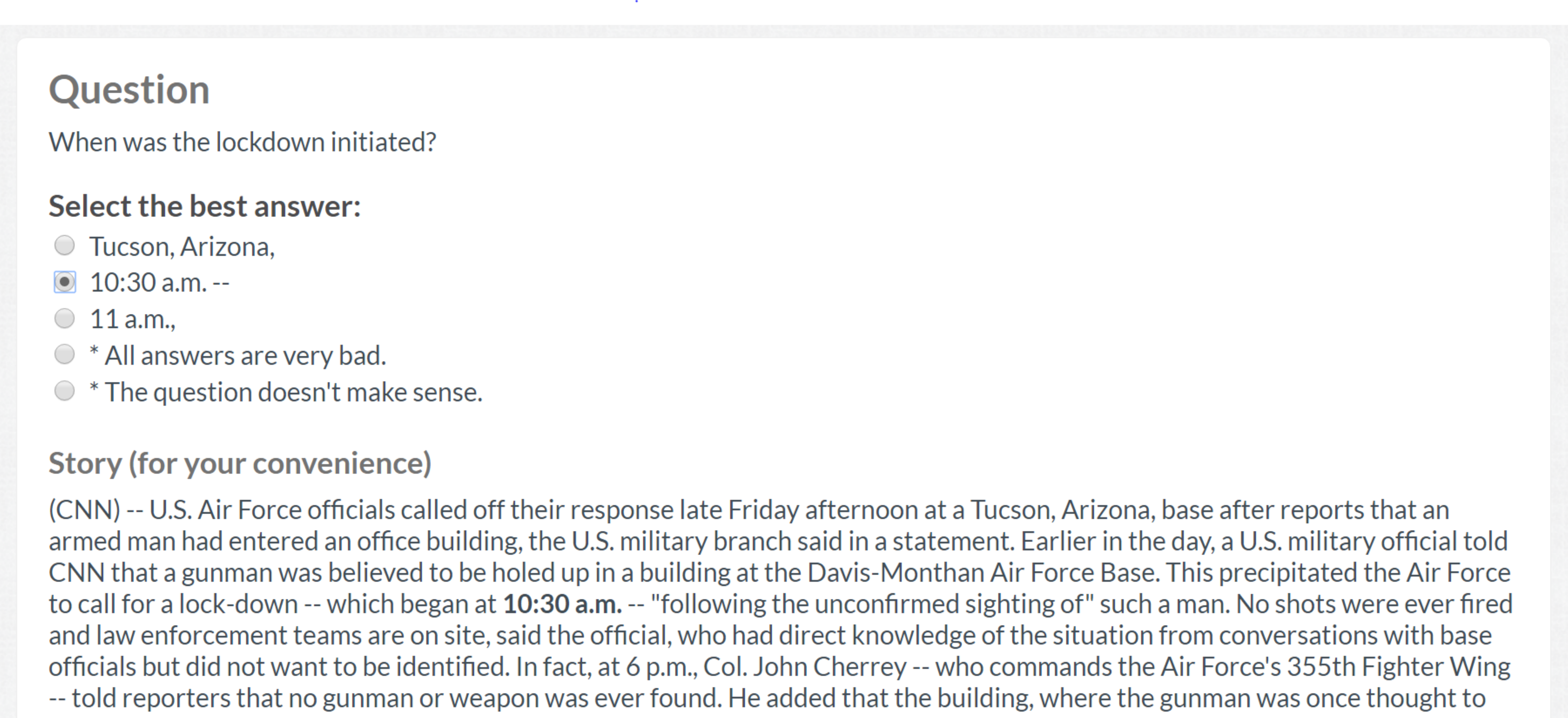}
	\caption{Examples of user interfaces for question sourcing, answer sourcing, and validation.}
	\label{fig:turk-ui}
\end{figure*}

\end{document}